\pdfoutput=1

\documentclass[11pt]{article}
\usepackage{multirow}
\usepackage{colortbl}
\usepackage{enumitem}
\usepackage{float}
\usepackage{xcolor}
\usepackage[final]{acl}
 \usepackage{subfig}
\usepackage{times}
\usepackage{latexsym}
\usepackage{amsmath}  
\usepackage{amsfonts} 
\usepackage{booktabs}
\usepackage{amssymb}  
\usepackage[T1]{fontenc}

\usepackage[utf8]{inputenc}
\usepackage{xcolor}
\usepackage{tcolorbox}

\usepackage{microtype}

\usepackage{inconsolata}

\usepackage{graphicx}
\newcommand\crule[3][black]{\textcolor{#1}{\rule{#2}{#3}}}
\definecolor{vio}{rgb}{0.4588235294117647, 0.4392156862745098, 0.7019607843137254}
\definecolor{grn}{rgb}{0.10588235294117647, 0.6196078431372549, 0.4666666666666667}
\usepackage[prependcaption,textsize=tiny]{todonotes}

\title{CAPE: Context-Aware Personality Evaluation Framework for Large Language Models}

\author{Jivnesh Sandhan, Fei Cheng, Tushar Sandhan\textsuperscript{$\dagger$} and Yugo Murawaki  \\
Kyoto University, Japan and \textsuperscript{$\dagger$}IIT Kanpur, India\\
\texttt{\{jivnesh,feicheng,murawaki\}@i.kyoto-u.ac.jp, \textsuperscript{$\dagger$}sandhan@iitk.ac.in}
\texttt{}}

\begin{document}
\maketitle
\begin{abstract}
Psychometric tests, traditionally used to assess humans, are now being applied to Large Language Models (LLMs) to evaluate their behavioral traits. However, existing studies follow a context-free approach, answering each question in isolation to avoid contextual influence. We term this the Disney World test, an artificial setting that ignores real-world applications, where conversational history shapes responses.

To bridge this gap, we propose the first Context-Aware Personality Evaluation (CAPE) framework for LLMs, incorporating prior conversational interactions. To thoroughly analyze the influence of context, we introduce novel metrics to quantify the consistency of LLM responses, a fundamental trait in human behavior. 

Our exhaustive experiments on 7 LLMs reveal that conversational history enhances response consistency via in-context learning but also induces personality shifts, with \texttt{GPT-3.5-Turbo} and \texttt{GPT-4-Turbo} exhibiting extreme deviations. While \texttt{GPT} models are robust to question ordering, \texttt{Gemini-1.5-Flash} and \texttt{Llama-8B} display significant sensitivity. Moreover, \texttt{GPT} models response stem from their intrinsic personality traits as well as prior interactions, whereas \texttt{Gemini-1.5-Flash} and \texttt{Llama--8B} heavily depend on prior interactions. Finally, applying our framework to Role Playing Agents (RPAs) shows context-dependent personality shifts improve response consistency and better align with human judgments.\footnote{Our code and datasets are publicly available at: \url{https://github.com/jivnesh/CAPE}}
\end{abstract}

\section{Introduction}
Large Language Models (LLMs) have made significant advances in generating human-like text \cite{spangher-etal-2024-llms,ou-etal-2024-dialogbench}. Moving beyond linguistic fluency, this raises the fundamental question: \textit{``How human-like are LLMs?''} Researchers are utilizing psychometrics to assess whether LLMs exhibit human-like personality traits \cite{Huang2024OnTH,serapiogarcía2023personalitytraitslargelanguage}. This question is crucial in contexts where LLMs may act as human proxies in surveys \cite{Dillion2023-lp,Harding2024-cf} or personalized human-AI interactions \cite{tseng-etal-2024-two}.
\begin{figure}[t]
\centering
    \centerline{\includegraphics[width=2.7in]{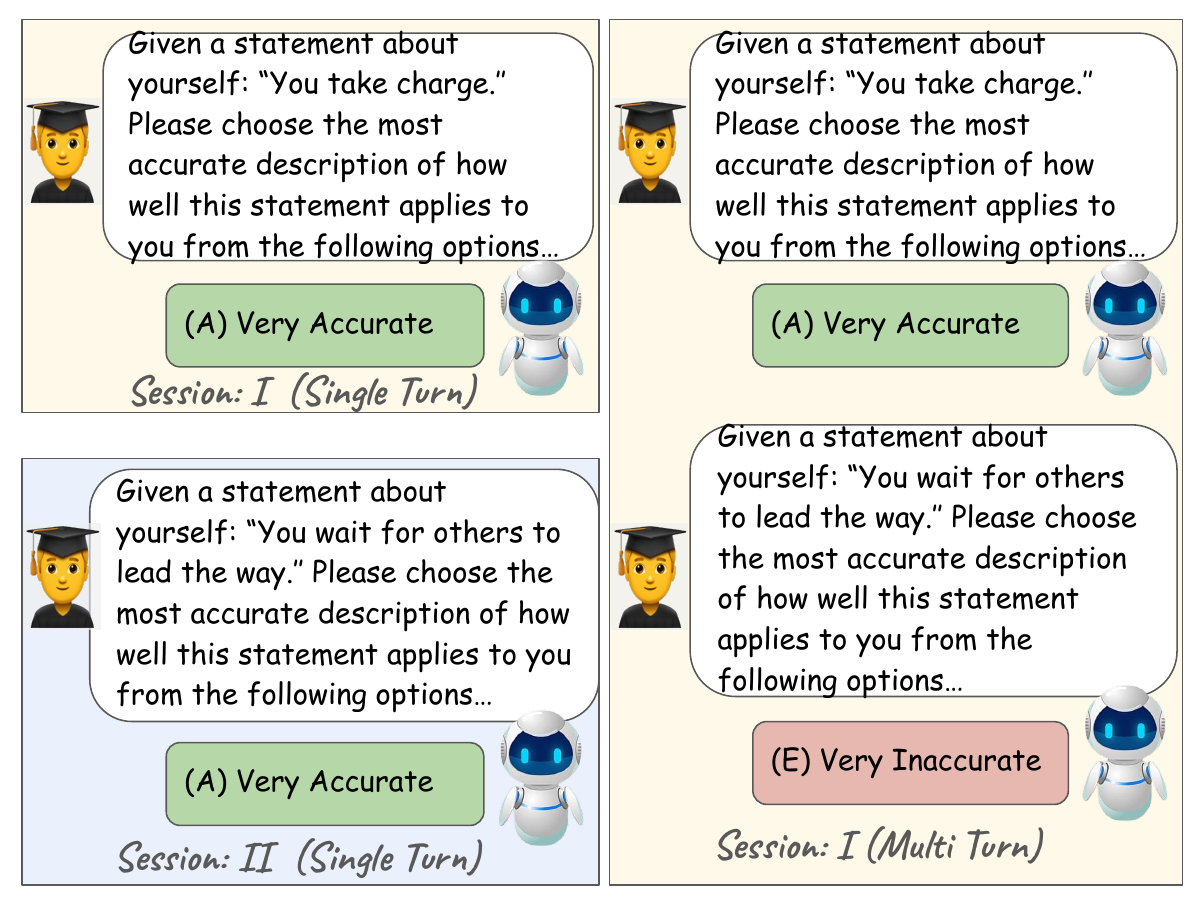}}
    \caption{Illustration of the difference between existing context-free (left) and the proposed context-aware (right) evaluation framework. In the left setting, each question is asked independently, with no prior history (indicated by a different background color). In the right setting, all questions are part of the same session, where prior answers can influence future responses.}
\label{fig:multi-turn} 
\end{figure}

Psychology provides well-established personality assessment frameworks, such as the Big Five Personality model 
\cite{McCrae1992-sk}, which evaluate individuals through a series of questions rated on a Likert scale. Analogously, LLMs are assessed using a zero-shot multiple-choice question (MCQ) format (\S \ref{preliminaries}). However, the reliability of these assessments remains a subject of debate. Some researchers advocate for methodologies that measure intrinsic personality traits in LLMs \cite{neurips23-spotlight,wang-etal-2024-incharacter,yang-etal-2023-psycot,jiang-etal-2024-personallm}, while others highlight inconsistencies stemming from prompt sensitivity \cite{shu-etal-2024-dont,gupta-etal-2024-self,song2023largelanguagemodelsdeveloped}.

We identify a critical research gap in existing research: LLM personality assessments are conducted in an isolated, context-independent manner. We term this the \textit{Disney World} test setting, where each question is answered without influence from prior responses. In contrast, real-world applications of LLMs necessitate exposure to conversational history. 
Before deploying LLMs in critical domains such as education and healthcare, it's vital to understand how conversational history impacts personality assessments.

To bridge this gap,  we propose the first Context-Aware Personality Evaluation (CAPE) framework, where prior questions and responses are retained in the conversational history to evaluate their impact on LLM personality (\S \ref{consistncy_framework}).
To thoroughly analyze the influence of context, we assess response consistency, a fundamental human trait, by introducing various inconsistency factors related to prompt sensitivity such as temperature, option wording, option order, instruction and item paraphrasing. To quantify this consistency, we present novel metrics that measure the response pattern similarity across multiple LLM runs (\S \ref{proposed_metrics}).

Our key findings on the impact of context on LLM personality are: Prior conversational history enhances response consistency by serving as few-shot in-context learning (\S \ref{why}). However, introducing context leads to deviations in responses compared to context-independent setting, with \texttt{GPT-3.5-Turbo} and \texttt{GPT-4-Turbo} exhibiting extreme personality shifts (\S \ref{how}). Additionally, \texttt{GPT} models maintain intrinsic personality despite contextual influence, whereas \texttt{Gemini-1.5-Flash} and \texttt{Llama-3.1-8B} rely heavily on prior conversation (\S \ref{where}). While these \texttt{GPT} models remain robust to question ordering, \texttt{Gemini-1.5-Flash} and \texttt{Llama-3.1-8B} display significant sensitivity (\S \ref{order-effect}).  Finally, we demonstrate our framework on Role Playing Agents (RPAs) and show context-dependent personality shifts improve response consistency and better align with human scores (\S \ref{human_alignemnt}). Our key contributions are:
\begin{itemize}
    \itemsep0em 
    \item To the best of our knowledge, we introduce the first Context-Aware Personality Evaluation (CAPE) framework, demonstrating its role in enhancing consistency (\S \ref{proposed_framework}).
    \item We propose novel metrics to quantify the consistency of LLM in assessments (\S \ref{proposed_metrics}).
    \item We conduct an in-depth analysis of how context influences LLM's personality (\S \ref{analysis}).
    
\end{itemize}

\section{Preliminaries: LLM's Personality Test}
\label{preliminaries}
The Big Five personality framework \cite{McCrae1992-sk,zis-JohnSrivastava1999The} characterizes human personality using 5 fundamental traits: \textbf{O}penness (artistic, imaginative), \textbf{C}onscientiousness (organized, thorough), \textbf{E}xtraversion (assertive, talkative), \textbf{A}greeableness (appreciative, kind), and \textbf{N}euroticism (anxious, worrying), collectively known as OCEAN. Following earlier works \cite{huang-etal-2024-reliability,neurips23-spotlight,zhou-etal-2023-realbehavior}, we assess the personality of an LLM by formulating the evaluation as a zero-shot multiple-choice question-answering task. 
Each assessment item consists of a self-descriptive statement and a set of response options. The model is prompted to evaluate how accurately the statement aligns with its personality by selecting the most appropriate response. The prompt for it as:  
   \begin{quote}
\textit{Given a statement about yourself: ``You \{Item\}.'' Please select the most accurate description of how well this statement applies to you from these options:}
\begin{enumerate}[label=(\Alph*)]
    \itemsep0em 
    \item \textit{Very Accurate}
    \item \textit{Moderately Accurate}
    \item \textit{Neither Accurate Nor Inaccurate}
    \item \textit{Moderately Inaccurate}
    \item \textit{Very Inaccurate}
\end{enumerate}
\end{quote} 
    where the \texttt{Item} describes behavioral tendency from a second-person perspective.  Each item corresponds to one of the 5 OCEAN dimensions and is either positively (+Key) or negatively (-Key) related to that dimension.  For example,  ($-O$): \textit{``Do not like poetry''} and ($+O$): \textit{``Love to daydream''} are negatively and positively correlated to openness, respectively.
    
The responses are numerically scored based on their alignment with the corresponding trait dimension. If an item is positively correlated, options (A) to (E) are scored from 5 to 1; otherwise, if negatively correlated, they are scored from 1 to 5.
For an assessment consisting of \( m \) items, the sequence of scores assigned to an LLM’s responses forms a \textbf{scoring trajectories}, which serve as the foundation for consistency analysis in our proposed framework. Mathematically, we define the scoring trajectory as, 
    $\mathcal{T} = [s_1, s_2, \dots, s_m]$
where   \( \mathcal{T} \) is the scoring trajectory of the LLM,  \( s_i \) is the score assigned to the LLM’s response for item \( i \) and \( m \) is the total number of items. This trajectory captures the model’s response pattern across all items and is later utilized in our framework to analyze the consistency of LLM-generated personality.

For a given OCEAN trait \( d \) (where \( d \in \{O, C, E, A, N\} \)), the trait score of an LLM is computed as the average score across all items associated with that dimension:
$S_d = \frac{1}{N_d} \sum_{i=1}^{N_d} s_i$,
where \( S_d \) is the OCEAN score for trait \( d \),  \( N_d \) is the total number of items related to trait \( d \), \( s_i \) is the score assigned to the LLM’s response for item \( i \). The final \textbf{OCEAN score} of an LLM consists of the 5 computed scores \( (S_O, S_C, S_E, S_A, S_N) \), which provide a quantitative representation of the model’s personality tendencies. We also consider all permutations of OCEAN to make its associated trajectory order invariant (\S \ref{proposed_metrics}). These trajectories are then analyzed for consistency and alignment.

\section{The Proposed Framework:  Context-Aware Personality Evaluation}
\label{proposed_framework}
In our proposed framework, we introduce a Context-Aware  Personality Evaluation (CAPE), where prior questions and responses are retained in the conversational history. To analyze the impact of context, we evaluate response consistency, a fundamental human trait, by considering various inconsistency factors related to prompt sensitivity, such as temperature, option wording, option order, instructions, and item paraphrasing. Finally, we introduce novel metrics to quantify this response consistency by measuring the similarity of response patterns across multiple LLM runs.  We recognize that ``context'' can indeed encompass various dimensions, including interlocutor attributes or external databases, as indicated in broader NLP research. However, our study specifically focuses on the influence of conversational history on personality assessments in LLMs, since our primary objective is to capture the influence of conversational history. 

\subsection{Context-Aware Evaluation}
\label{consistncy_framework}
Traditional LLM personality assessments treat responses in isolation, ignoring prior interactions. However, real-world applications involve multi-turn conversations where the context shapes responses. 
To bridge this gap, we propose a context-dependent personality assessment framework that retains prior exchanges while answering new questions (Figure \ref{fig:multi-turn}). This approach enables a more \textit{realistic} evaluation by assessing whether LLMs maintain consistent personality traits or shift based on context—critical for applications like AI tutoring, virtual assistants, and social chatbots.
Formally, let \( Q = \{q_1, q_2, \dots, q_m\} \) be a set of personality assessment questions and \( R = \{r_1, r_2, \dots, r_m\} \) the corresponding LLM responses. At any time step \( t \), the conversational history is defined as \( H_{t-1} = \{(q_1, r_1), (q_2, r_2), ..., (q_{t-1}, r_{t-1})\} \), and the LLM’s response function becomes \( r_t = f(q_t, H_{t-1}) \), incorporating prior interactions.

\paragraph{Inconsistency Factors}
\label{inconsistency_factors}
We introduce 5 sensitivity factors: temperature, option wording, option order, instruction, and item paraphrasing—each with 3 variants. For each variant, we generate 3 scoring trajectories using the same question order across independent LLM runs. We assess response consistency with multiple metrics, with temperature set to 0 except when testing its sensitivity. Refer to Appendix \ref{sec:prompt_templates} for detailed prompts and examples.\\
\textbf{(1) Stability:} We establish a baseline by running the assessment 3 times without sensitivity factors, providing a reference for their impact.\\
\textbf{(2) Temperature:} We test 3 temperature values: $0.5, 1,$ and $1.5$. While prior studies commonly use a default temperature of $0$, real-world applications may require non-zero temperatures for more dynamic and adaptable behavior \cite{lee-etal-2025-evaluating}.\\
\textbf{(3) Option Wording:}  We experiment with 3 paraphrased versions of each option while maintaining semantic equivalence \cite{shu-etal-2024-dont}. For example, ``\textit{Strongly agree}'' is reworded as ``\textit{Completely Aligned}'' and ``\textit{Perfectly Compatible.}''\\
\textbf{(4) Option Order:} We explore three ordering variations: original order \textit{(A B C D E)}, reverse order \textit{(E D C B A)}, and a randomized order \textit{(C B D A E)} \cite{gupta-etal-2024-self,song2023largelanguagemodelsdeveloped}.\\
\textbf{(5) Instruction:} Previous studies have used different instruction formulations for personality evaluation \cite{huang-etal-2024-reliability,neurips23-spotlight,serapiogarcía2023personalitytraitslargelanguage}. We assess consistency across 3 variations of the instruction prompt.\\
\textbf{(6) Item Paraphrasing:} Using \texttt{GPT4}, we generate 2 paraphrased versions of each item \cite{huang-etal-2024-reliability}. For example, the original item \textit{``Worry about things''} is reworded as \textit{``Have anxiety about situations''} and \textit{``Stress over issues.''} We manually verify paraphrased items for semantic fidelity to preserve its validity and reliability.

\subsection{The Proposed Consistency Metrics}
\label{proposed_metrics}
Standard consistency metrics that measure exact
response agreement \cite{atil2024llmstabilitydetailedanalysis} or Euclidean distance between
trajectories overlook partial agreement and contextual dependencies between scores, treating all divergences equally. In response, we propose 2 novel metrics: Trajectory Consistency (TC) and OCEAN Consistency (OC). Our metrics focus on capturing the similarity in patterns of responses across multiple runs. By applying Gaussian Process Regression (GPR) independently on each trajectory, we account for both the responses and their interactions with neighboring questions. Our metrics assess consistency by evaluating the ratio of the intersection to the union of the posterior predictive distribution's support at each point. A higher consistency corresponds to a greater overlap (intersection) of confidence intervals, while increased inconsistency results in a larger union (wider spread) of these intervals. This proportional relationship provides a clear measure of how consistent the model's responses are across different assessment runs.

\paragraph{Trajectory Consistency (TC):} 
Trajectory consistency refers to the similarity between three scoring trajectories produced when the same LLM takes the same psychometric test three times. Inconsistencies often arise due to inherent stochasticity in LLM outputs, sensitivity to prompt variations (e.g., option wording, order, or instructions), and lack of contextual grounding—all of which can cause response shifts despite an unchanged assessment. This is problematic because, like humans, a stable personality should yield consistent answers across repeated assessments; fluctuations undermine the reliability and interpretability of model behavior. Each LLM is assessed three times, producing three scoring trajectories. A trajectory is considered more consistent when these are closer in pattern and distance, which we quantify using our proposed Trajectory Consistency metric. A higher score indicates greater response stability.

Each scoring trajectory \( \mathcal{T}_i \) represents an independent run of the LLM's personality assessment, where \( i \in \{1,2,3\} \). Each trajectory consists of pairs \( (x_t, y_{i,t}) \) for \( t \in \{1, \dots, m\} \), where \( x_t \) is the index of \( t \)-th question, and \( y_{i,t} \) is the score assigned in the \( i \)-th run.
We apply a moving average filter with a window size $\omega$\footnote{We use $\omega = 4$ based on our hyper-parameter tuning.} for outlier denoising: $\hat{y}_{i,t} = \frac{1}{\omega}\sum_{j=0}^{\omega-1} y_{i,t-1}$, followed by mean normalization: $
\hat{y}_{i,t} = \frac{\hat{y}_{i,t} - \mu_i}{\sigma_i}$
where \( \mu_i \) and \( \sigma_i \) are the mean and standard deviation of the corresponding smoothed trajectory.
Then, we model each normalized trajectory using GPR as $
f_i(x) \sim \mathcal{GP}(\mu_i(x), k_i(x, x')),$ where $\mu_i(x), k_i(x, x')$ are mean and kernel function.\footnote{We use the Radial Basis Function kernel and automate hyper-parameter tuning with the scikit-learn library.}  It gives the posterior predictive distribution at each \( x_t \) as $
f_i(x_t) \sim \mathcal{N}(\mu_i(x_t), \sigma_i^2(x_t)),$ where $\mu_i(x_t), \sigma_i^2(x_t)$ are posterior mean and variance respectively.
We define \textit{support interval} as $
S_i(x_t) = [L_i(x_t), U_i(x_t)].$\footnote{ For example,  for the 95\% confidence region, $
L_i(x_t) = \mu_i(x_t) - 1.96\sigma_i(x_t), \quad U_i(x_t) = \mu_i(x_t) + 1.96\sigma_i(x_t).$}
The intersection of all $3$ supports at \( x_t \) is 
\begin{equation}
\small
    W_{\text{int}}(x_t) = \max \left( 0, \min_i U_i(x_t) - \max_i L_i(x_t) \right)\end{equation}
The union of all supports is computed by first sorting the support intervals in ascending order based on their lower bounds and then iteratively merging the overlapping intervals followed by summing the lengths of all merged, non-overlapping supports as 

\begin{equation}
\small
   W_{\text{union}}(x_t) = \sum_{j} \left( U_j^{\text{merged}}(x_t) - L_j^{\text{merged}}(x_t) \right)
\end{equation}

where \( U_j^{\text{merged}}(x_t) \) and \( L_j^{\text{merged}}(x_t) \) represent the upper and lower bounds of the merged segments.
Finally, consistency score $TC$ is calculated as
\begin{equation}
\small
TC = \frac{1}{x_{m}} \int_{0}^{x_{m}} \frac{W_{\text{int}}(x_t)}{W_{\text{union}}(x_t)} \, dx.
\end{equation}

\paragraph{OCEAN Consistency (OC):}
We obtain an OCEAN score from each scoring trajectory (\S \ref{preliminaries}) and write it as:  $
\mathbf{s}_i = (O_i, C_i, E_i, A_i, N_i),\quad i \in \{1,2,3\}$
We generate all possible orderings of the 5 traits to make this representation order invariable as $
\mathcal{P}(\mathbf{s}_i) = \{ \pi_j(\mathbf{s}_i) \}_{j=1}^{5!} = \{ \mathbf{s}_{i,j} \}_{j=1}^{120}.$ Then, we build a time series by appending each permuted sequence \( \mathbf{s}_{i,j} \) for the \( i \)-th trajectory as:  $
\mathcal{T}_{i} = \{ (x_t, y_{i,j,t}) \}_{t=1}^{5 \times 120}$, where \( y_{i,j,t} \) represents the score at position \( t \) in the permuted sequence \( \mathbf{s}_{i,j} \).  
This transformation makes the series representation order-invariant. Then, we plug this series in the above formulation to obtain the consistency score. We call this as OCEAN consistency score.

\section{Experimental Setup}
\label{exp-setup}
\paragraph{Datasets:} 
We use 2 datasets: the Machine Personality Inventory (MPI) \cite{neurips23-spotlight} (\S \ref{result-section}), licensed under MIT, which includes 120 items from the International Personality Item Pool (IPIP) and its IPIP-NEO adaptations \cite{Goldberg1999ABP,McCrae1997-vk}, and the Big Five Inventory (BFI) (\S \ref{human_alignemnt}) with 44 items \cite{Lang2011-fk}.  

\paragraph{Construct validity and reliability of the psychometric instruments used:}
We would like to clarify that our work does not introduce new psychometric instruments, but rather builds on top of well-established ones in a more realistic, context-dependent evaluation setting. The validity and reliability of these instruments—such as the IPIP and BFI—have already been demonstrated in prior work on LLMs under context-independent settings. For example, \newcite{serapiogarcía2023personalitytraitslargelanguage} conducted a large-scale study across 18 LLMs showing strong construct validity and reliability of psychometric assessments. Similarly, \newcite{Wang2025} reported high convergent validity between LLM-based and human-reported personality scores. Moreover, a growing body of research \cite{neurips23-spotlight,wang-etal-2024-incharacter,yang-etal-2023-psycot,jiang-etal-2024-personallm} has consistently applied these psychometric instruments to LLMs, further reinforcing their validity and reliability. Our framework builds directly on these validated instruments, differing only in that it adopts a human-like interaction setting where conversational history is retained. 
\begin{table*}[h]
\centering
\resizebox{0.7\textwidth}{!}{%
\begin{tabular}{cc|cccc|cccc}
\hline
\multicolumn{1}{l}{}                  & \multicolumn{1}{l}{} & \multicolumn{4}{c}{context-free} & \multicolumn{4}{c}{context-dependent} \\
\cmidrule(r){1-1}\cmidrule(r){2-2}\cmidrule(r){3-6}\cmidrule(r){7-10}
LLM system                            & Sensitivity factors  & TAR$(\uparrow)$     & ED$(\downarrow)$   & TC$(\uparrow)$      & OC$(\uparrow)$      & TAR$(\uparrow)$     & ED$(\downarrow)$  & TC$(\uparrow)$     & OC$(\uparrow)$     \\
\cmidrule(r){1-1}\cmidrule(r){2-2} \cmidrule(r){3-6}\cmidrule(r){7-10}
\multirow{6}{*}{\texttt{GPT-3.5-Turbo}}        & Stability            & 86.67   & 0.16  & 28.34   & 77.88   & \cellcolor{green!10}91.67   & \cellcolor{green!10}0.08 & \cellcolor{green!10}72.89  & \cellcolor{green!10}91.46  \\
                                      & Temperature          & 40.83   & 0.76  & 38.60   & 63.67   & \cellcolor{green!10}71.67   &\cellcolor{green!10} 0.26 \cellcolor{green!10}& \cellcolor{green!10}47.26  & \cellcolor{green!10}85.25  \\
                                      & Option Wording       & 11.67   & 0.71  & 39.08   & 67.04   & \cellcolor{green!10}70.00   & \cellcolor{green!10}0.28 & \cellcolor{green!10}47.94  & \cellcolor{green!10}87.10  \\
                                      & Option order         & 11.67   & 0.78  & 32.82   & 63.48   & \cellcolor{green!10}59.17   &\cellcolor{green!10} 0.33 & \cellcolor{green!10}39.30  & \cellcolor{green!10}83.26  \\
                                      & Instructions         & \cellcolor{green!10}25.83   & 0.95  & 17.74   & 66.82   & 20.83   & \cellcolor{green!10}0.74 & \cellcolor{green!10}34.71  & \cellcolor{green!10}71.50  \\
                                      & Item paraphrasing    & \cellcolor{green!10}71.67   & \cellcolor{green!10}0.36  & \cellcolor{green!10}34.74   & 68.64   & 60.00   & 0.41 & 33.50  & \cellcolor{green!10}77.28  \\
                                      \cmidrule(r){2-2} \cmidrule(r){3-6}\cmidrule(r){7-10}
\multirow{6}{*}{\texttt{GPT-4-Turbo}}          & Stability            & 84.17   & 0.23  & 39.50   & \cellcolor{green!10}90.62   & \cellcolor{green!10}92.50   & \cellcolor{green!10}0.17 & \cellcolor{green!10}69.31  & 90.24  \\
                                      & Temperature          & 70.83   & 0.39  & 49.15   & 83.74   & \cellcolor{green!10}90.00   & \cellcolor{green!10}0.12 & \cellcolor{green!10}76.52  & \cellcolor{green!10}91.62  \\
                                      & Option Wording       & 69.17   & 0.40  & 33.92   & 81.65   & \cellcolor{green!10}92.50   & \cellcolor{green!10}0.16 & \cellcolor{green!10}65.54  & \cellcolor{green!10}88.87  \\
                                      & Option order         & 45.00   & 0.99  & 17.33   & 72.00   & \cellcolor{green!10}90.83   & \cellcolor{green!10}0.18 & \cellcolor{green!10}68.30  & \cellcolor{green!10}90.45  \\
                                      & Instructions         & \cellcolor{green!10}52.50   & \cellcolor{green!10}0.63  & 23.65   & \cellcolor{green!10}75.70   & 32.50   & 0.97 & \cellcolor{green!10}32.90  & 71.31  \\
                                      & Item paraphrasing    & 54.17   & 0.62  & 28.59   & \cellcolor{green!10}82.48   & \cellcolor{green!10}85.00   & \cellcolor{green!10}0.34 & \cellcolor{green!10}50.73  & 80.29  \\
                                      \cmidrule(r){2-2} \cmidrule(r){3-6}\cmidrule(r){7-10}
\multirow{6}{*}{\texttt{Gemini-1.5-Flash}}     & Stability            & \cellcolor{green!10}100.00  & \cellcolor{green!10}0.00  & \cellcolor{green!10}100.00  & \cellcolor{green!10}100.00  & \cellcolor{green!10}100.00  & \cellcolor{green!10}0.00 & \cellcolor{green!10}100.00 & \cellcolor{green!10}100.00 \\
                                      & Temperature          & 90.00   & 0.08  & 71.56   & \cellcolor{green!10}92.93   & \cellcolor{green!10}93.33   & \cellcolor{green!10}0.07 & \cellcolor{green!10}74.86  & 88.66  \\
                                      & Option Wording       & 59.17   & 0.35  & 45.29   & 76.24   & \cellcolor{green!10}75.83   &\cellcolor{green!10} 0.21 & \cellcolor{green!10}50.67  & \cellcolor{green!10}82.07  \\
                                      & Option order         & 36.67   & 0.57  & 30.90   & 76.56   & \cellcolor{green!10}40.83   & \cellcolor{green!10}0.47 & \cellcolor{green!10}42.01  & \cellcolor{green!10}78.91  \\
                                      & Instructions         & 45.83   & 0.48  & \cellcolor{green!10}35.01   & \cellcolor{green!10}78.30   & \cellcolor{green!10}65.83   & \cellcolor{green!10}0.38 & 31.39  & 71.36  \\
                                      & Item paraphrasing    & 57.50   & 0.38  & 33.98   & \cellcolor{green!10}78.75   & \cellcolor{green!10}65.00   & \cellcolor{green!10}0.37 & \cellcolor{green!10}34.76  & 70.43  \\
                                     \cmidrule(r){2-2} \cmidrule(r){3-6}\cmidrule(r){7-10}
\multirow{6}{*}{\texttt{Claude-3-5-Haiku}} & Stability         & \cellcolor{green!10}100.00 &\cellcolor{green!10} 0.00 & \cellcolor{green!10}100.00 & \cellcolor{green!10}100.00 &\cellcolor{green!10} 100.00 & \cellcolor{green!10}0.00 &\cellcolor{green!10} 100.00 & \cellcolor{green!10} 100.00 \\
&Temperature       &\cellcolor{green!10} 71.67  &\cellcolor{green!10} 0.23 & \cellcolor{green!10} 51.57   & \cellcolor{green!10}85.31 & 67.50  & 0.24 & 32.86  &  76.28   \\
&Option Wording    & 25.00  & 0.88 & \cellcolor{green!10} 19.96   & \cellcolor{green!10} 68.03  &\cellcolor{green!10} 34.17  &\cellcolor{green!10} 0.49 &18.18   & 67.61  \\
&Option order      & 24.17  & 1.04 & 17.35  & \cellcolor{green!10}74.02  &\cellcolor{green!10} 43.33  &\cellcolor{green!10} 0.49 &  \cellcolor{green!10} 17.71  &  63.96  \\
&Instructions      & 25.83  & 0.74 & \cellcolor{green!10}23.17 &  \cellcolor{green!10}77.35 &\cellcolor{green!10} 32.50  & \cellcolor{green!10}0.55 &  8.21   &  67.49  \\
&Item paraphrasing & 28.33  & 0.84 & \cellcolor{green!10}19.53  & \cellcolor{green!10} 83.21  &\cellcolor{green!10} 65.00  & \cellcolor{green!10}0.26 & 19.17  & 66.19  \\
\cmidrule(r){2-2} \cmidrule(r){3-6}\cmidrule(r){7-10}
\multirow{6}{*}{\texttt{Llama-3.1-8B}} & Stability            & \cellcolor{green!10}100.00  & \cellcolor{green!10}0.00  & \cellcolor{green!10}100.00  & \cellcolor{green!10}100.00  & \cellcolor{green!10}100.00  & \cellcolor{green!10}0.00 & \cellcolor{green!10}100.00 & \cellcolor{green!10}100.00 \\
                                      & Temperature          & \cellcolor{green!10}40.00   & \cellcolor{green!10}0.76  & \cellcolor{green!10}30.13   & \cellcolor{green!10}76.73   & 24.17   & 1.00 & 25.75  & 67.59  \\
                                      & Option Wording       & 39.17   & 0.73  & 34.32   & \cellcolor{green!10}84.74   & \cellcolor{green!10}45.83   & \cellcolor{green!10}0.51 & \cellcolor{green!10}38.92  & 83.32  \\
                                      & Option order         & 8.33    & 1.17  & 17.67   & 71.23   & \cellcolor{green!10}25.83   & \cellcolor{green!10}0.68 & \cellcolor{green!10}23.48  & \cellcolor{green!10}73.43  \\
                                      & Instructions         & \cellcolor{green!10}39.17   & \cellcolor{green!10}0.76  & 30.71   & \cellcolor{green!10}79.37   & 16.67   & 0.81 & \cellcolor{green!10}38.34  & 74.26  \\
                                      & Item paraphrasing    & \cellcolor{green!10}55.00   & 0.54  & 38.37   & \cellcolor{green!10}85.87   & 50.83   & \cellcolor{green!10}0.52 & \cellcolor{green!10}41.64  & 78.62 \\
                                       \cmidrule(r){2-2} \cmidrule(r){3-6}\cmidrule(r){7-10}
\multirow{6}{*}{\texttt{Llama-3.3-70B}} &Stability         & 99.17  & 0.01 & 96.34  & 98.08  & \cellcolor{green!10}100.00 & \cellcolor{green!10}0.00 & \cellcolor{green!10}100.00 & \cellcolor{green!10}100.00 \\
&Temperature       & 88.33  & 0.11 & 67.19  & 83.24  &\cellcolor{green!10} 94.17  &\cellcolor{green!10} 0.05 &\cellcolor{green!10} 85.56  & \cellcolor{green!10} 92.59  \\
&Option Wording    & 49.17  & 0.44 & 42.03  & 76.33  & \cellcolor{green!10}83.33  &\cellcolor{green!10} 0.14 &\cellcolor{green!10} 65.38  & \cellcolor{green!10}87.25  \\
&Option order      & 32.50  & 0.63 & 45.50  & 78.51  & \cellcolor{green!10}35.83  &\cellcolor{green!10} 0.47 &\cellcolor{green!10} 57.83  & \cellcolor{green!10}79.19  \\
&Instructions      & 13.33  & 1.03 & 23.54  & 66.99  & \cellcolor{green!10}76.67  & \cellcolor{green!10}0.17 &\cellcolor{green!10} 62.85  & \cellcolor{green!10}89.66  \\
&Item paraphrasing & 46.67  & 0.50 & 39.11  & 75.58  & \cellcolor{green!10}73.33  & \cellcolor{green!10}0.26 &\cellcolor{green!10} 38.97  & \cellcolor{green!10}83.06  \\
\cmidrule(r){2-2} \cmidrule(r){3-6}\cmidrule(r){7-10}

\multirow{6}{*}{\texttt{Llama-3.1-405B}} &Stability         & 92.50  & 0.07 & 83.06  &\cellcolor{green!10} 96.81  & \cellcolor{green!10}95.83  & \cellcolor{green!10} 0.04 & \cellcolor{green!10} 85.92  & 95.41  \\
&Temperature       & 45.83  & 0.65 & 36.16  & 87.18  & \cellcolor{green!10} 57.50  &\cellcolor{green!10} 0.39 & \cellcolor{green!10} 45.11  &\cellcolor{green!10} 83.84  \\
&Option Wording    & \cellcolor{green!10}61.67  & 0.52 & 29.90  &\cellcolor{green!10} 87.65  & 36.67  &\cellcolor{green!10} 0.44 &\cellcolor{green!10} 52.53  & 84.71  \\
&Option order      & 54.17  & 0.61 & 41.83  & 84.78  &\cellcolor{green!10} 66.67  &\cellcolor{green!10} 0.22 &\cellcolor{green!10} 59.13  & \cellcolor{green!10} 93.10  \\
&Instructions      & 43.33  & 0.76 & 29.39  & 78.49  &\cellcolor{green!10} 43.33  &\cellcolor{green!10} 0.45 &\cellcolor{green!10} 46.97  & \cellcolor{green!10} 88.74  \\
&Item paraphrasing & 54.17  & 0.63 & 29.53  & 83.05  &\cellcolor{green!10} 63.33  &\cellcolor{green!10} 0.35 &\cellcolor{green!10} 36.35  & \cellcolor{green!10} 84.26 \\

                                      \hline
                                    
\end{tabular}}
\caption{Consistency evaluation on the MPI dataset under context-dependent and context-free settings across 5 different inconsistency factors. Each inconsistency factor has 3 variants, leading to 3 scoring trajectories. Consistency is measured using 4 metrics on these 3 trajectories. Higher values for Total Agreement Rate (TAR), Trajectory Consistency (TC), and OCEAN Consistency (OC) indicate better consistency, while lower values for the average pairwise Euclidean Distance (ED) are preferable. Each metric captures a distinct aspect of trajectory consistency and is not necessarily correlated with the others (\S \ref{exp-setup}). The best results for each row are highlighted in green. Overall, the context-dependent setting improves consistency, with notable variations across LLMs.}

\label{tab:main_results}
\end{table*}
\paragraph{Systems:} 
  To evaluate LLM personality and enhance the generalizability of our findings, we select 7 diverse LLMs that vary in architecture, alignment strategies, and model size:: \texttt{GPT-3.5-Turbo} \cite{openai2022gpt}, \texttt{GPT-4-Turbo} \cite{openai2024gpt4technicalreport}, \texttt{Gemini-1.5-Flash} \cite{team2024gemini}, \texttt{Claude-3.5-Haiku} \cite{claude} and \texttt{LLaMA-3.1-8B}, \texttt{Llama-3.3-70B}, \texttt{Llama-3.1-405B} \cite{grattafiori2024llama3herdmodels}. 

\paragraph{Evaluation Metrics}
We evaluate the following metrics on 3 scoring trajectories \( \mathcal{T}_i \), where \( i \in \{1,2,3\} \) defined over \( m \) time steps:
\begin{itemize}
\itemsep0em 
    \item \textbf{TAR($\uparrow$)}:
    The Total Agreement Rate (TAR) \cite{atil2024llmstabilitydetailedanalysis} measures the percentage of questions where the scores across 3 scoring trajectories are identical at each time step. $TAR$ is defined as $
\frac{1}{m} \sum_{k=1}^{m} \mathbf{1} (\mathcal{T}_{1,k} = \mathcal{T}_{2,k} = \mathcal{T}_{3,k}) 
$, where \( \mathbf{1}(\cdot) \) is the indicator function, which is 1 if the scores at time step \( k \) across all 3 trajectories are identical, and 0 otherwise. Higher TAR indicates higher consistency.
\item \textbf{ED($\downarrow$)}: The average pairwise Euclidean Distance (ED) measures divergence, with lower values indicating higher consistency.
\begin{equation}
   \small 
ED = \frac{1}{3m}  \sum_{k=1}^{m} \sum_{\substack{(i,j) \in \\ \left\{ (1,2), (1,3), (2,3) \right\}}} \| (\mathcal{T}_{i,k} - \mathcal{T}_{j,k})\|
\end{equation}
\item \textbf{TC($\uparrow$)} and \textbf{OC($\uparrow$):} Refer to \S \ref{proposed_metrics} for details. Higher values indicates higher consistency.

\end{itemize}
We use multiple metrics to ensure a comprehensive evaluation. Each metric captures a unique aspect of trajectory consistency and may not correlate with others. TAR evaluates exact pointwise agreement,  ED quantifies pairwise deviations per question and TC measures scoring pattern similarity. A model can have low TAR, high ED, yet high TC. Similarly, OC is sensitive to response of specific questions and may not align with other metrics.  
\section{Results}
\label{result-section}
Table \ref{tab:main_results} shows the results of 7 LLMs on the MPI dataset, evaluated under context-dependent and context-free settings across 5 inconsistency factors, each with 3 variations. We use 4 metrics to assess consistency, with the best results highlighted in green. Overall, context-dependent evaluation improves consistency, as shown by the green-marked values. In stability, \texttt{GPT-3.5-Turbo} and \texttt{GPT-4-Turbo} are inconsistent even at temperature 0, unlike \texttt{Gemini-1.5-Flash} and \texttt{Llama-3.1-8B}, but context-dependent evaluation improves their consistency.\footnote{OpenAI states: \textit{``Chat completions are non-deterministic.''} Models are inconsistent even with a fixed seed.} In the temperature factor, the context-dependent setting enhances consistency across all models except \texttt{Llama-3.1-8B}, which exhibits unstable trajectories and struggles to effectively leverage context, likely due to its smaller size. For option wording and option order, the context-dependent setting consistently outperforms the context-free setting. In terms of instruction sensitivity, \texttt{GPT-4-Turbo} shows reduced consistency in the context-dependent setting due to its heavy instruction tuning, which leads to deviations from semantically similar instructions \cite{zhou-etal-2024-paraphrase, Stribling2024}. Similarly, \texttt{Llama-3.1-8B} is also sensitive to instruction variations. Regarding item paraphrasing, \texttt{GPT-3.5-Turbo} does not perform better context-dependent, supporting prior research on its sensitivity to paraphrasing \cite{zhou-etal-2024-paraphrase, haller-etal-2024-yes}. \texttt{Llama-3.1-8B} shows similar performance in both settings. Among the \texttt{LLaMA} variants, larger models exhibit noticeably stronger consistency than their smaller counterparts (e.g., \texttt{LLaMA-3.1-8B)}, providing further support for our hypothesis that model size contributes to stable personality expression in context-rich settings. In summary, when the context-free setting outperforms the context-dependent setting, the model tends to be smaller in size with poor in-context learning ability and hypersensitive to the inconsistency factors.

\paragraph{Statistical Analysis of the Proposed Metrics:}
To empirically establish the statistical validity and robustness of our metrics, we conduct the following analysis (Refer to Appendix \ref{statistical_analysis} for details):\\
\textbf{Correlation Analysis:} We compute correlations of our metrics with baselines. Our results show strong positive correlations of TC (\textit{Pearson} $r = 0.77$, $p < 10^{-9}$; \textit{Spearman} $\rho = 0.76$, $p < 10^{-9}$) and OC ($r = 0.81$, $p < 10^{-11}$; $\rho = 0.79$, $p < 10^{-10}$) with TAR, and strong negative correlations with ED (TC: $r = -0.80$, $p < 10^{-10}$; OC: $r = -0.79$, $p < 10^{-10}$), reinforcing that TC and OC effectively measure consistency.\\
 \textbf{Reliability Analysis:} We evaluate metric stability through repeated trials, computing Cronbach’s alpha and test-retest reliability correlations. TC ($\alpha = 0.91$, test-retest $= 0.89$) and OC ($\alpha = 0.86$, test-retest $= 0.83$) demonstrate strong internal reliability and stability, surpassing or equaling established metrics (TAR and ED).\\
\textbf{Construct Validity:} We validate the metrics by applying them across context-dependent and context-free experimental conditions, observing significant differences for TC (\textit{ANOVA} $p = 0.0006$, Cohen’s $d = 0.81$) and OC ($p = 0.0084$, $d = 0.60$). These results confirm our metrics’ sensitivity to meaningful changes in experimental conditions.

\section{Analysis}
\label{analysis}
\subsection{What does make a trajectory consistent?}
\label{why}
To examine the role of context in consistency, we conduct an ablation study on \texttt{GPT-3.5-Turbo}, incrementally increasing the number of question-response pairs as few-shot demonstrations. Instead of preserving the full history, we keep only the most recent pairs, discarding older ones.
\begin{figure}[tbh]
\centering
    \centerline{\includegraphics[width=2.8in]{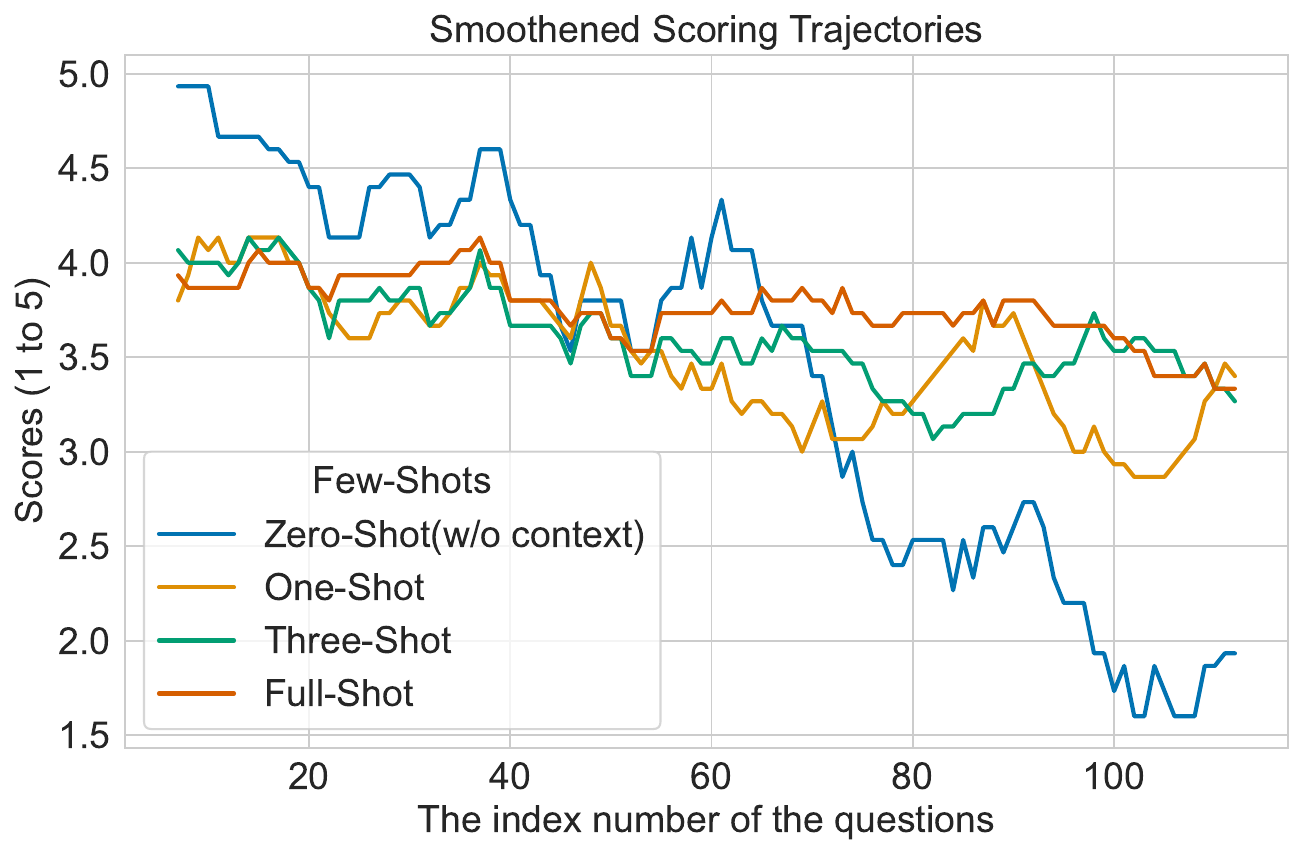}}
    \caption{This figure illustrates the mechanism behind the consistency, not the consistency itself. 
    Each trajectory corresponds to a different number of few shots; as the number increases, the trajectory approaches that of the full-shot context-dependent setting (Full-Shot: Red color).
    This indicates that prior pairs act as implicit few-shot demonstrations, enabling in-context learning.}
\label{fig:ablation} 
\end{figure}
Figure \ref{fig:ablation} illustrates the mechanism behind the consistency, not the consistency itself. It shows how in-context learning—retaining previous question-answer pairs (few-shots)—improves consistency. Each trajectory in Figure \ref{fig:ablation} corresponds to a different number of few-shots; as the number increases, the trajectory approaches that of the full-shot context-dependent setting. This illustrates how contextual grounding enhances consistency, as reported in Table \ref{tab:main_results}.
Figure \ref{fig:ablation} shows that as we increase the few-shots, response trajectories stabilize, eventually converging to the full-history setting (Red color). This suggests that prior question-response pairs act as implicit few-shot demonstrations, facilitating in-context learning.
Our results align with prior work showing that more in-context examples enhance consistency \cite{song-etal-2025-many,min-etal-2022-rethinking}.  While in-context learning enhances consistency, it does not guarantee logical consistency (\S \ref{sec:logical-consistency}).

\subsection{How does context affect LLM responses?}
\label{how}
\begin{figure}[!h]
\centering
\subfloat[\label{fig:trajectory_difference}]{\includegraphics[width=2.75in]{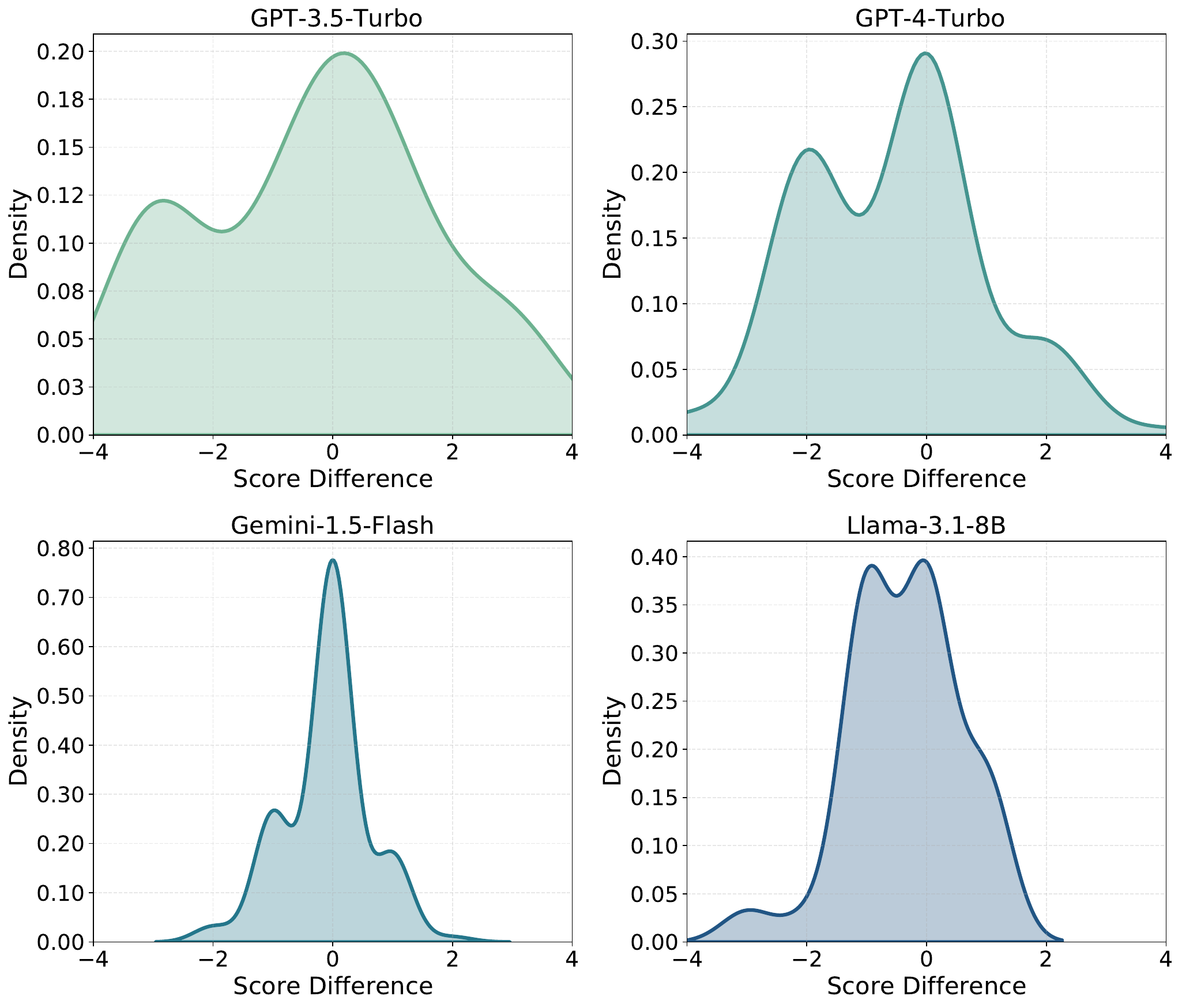}}\\
\subfloat[\label{fig:Ocean_plot}]{\includegraphics[width=2.75in]{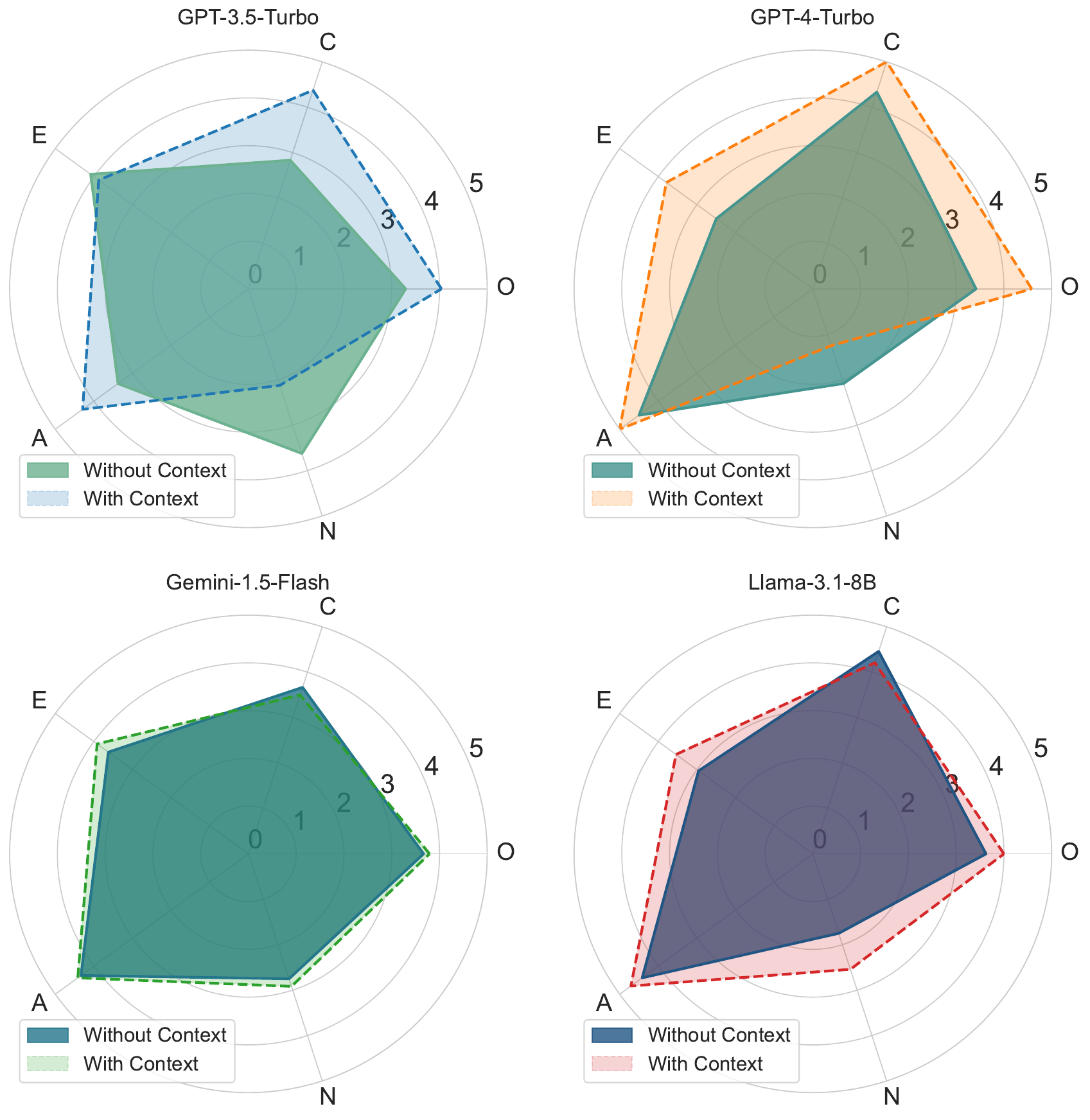}}
\caption{(a) Distribution indicates how frequently and to what extent the LLM alters its responses for the same item, if we simply switch from the context-independent setting to the context-dependent setting. A wider spread indicates greater deviation between these 2 settings. (b) Comparison of OCEAN personality traits for each LLM under both context-free (dark color) and context-dependent (light color) settings. \texttt{GPT-3.5-Turbo} and \texttt{GPT-4-Turbo} show significant personality shifts.} 
\label{fig:shift} 
\end{figure} 
\noindent We analyze the scoring trajectories of an LLM under both with/context-free settings, examining how frequently and to what extent the LLM alters its responses for the same item. To quantify these changes, we compute the pointwise score differences across all items and categorize them into discrete buckets ranging from -4 to 4. A difference near -4 or 4 signifies a complete polarity shift (e.g., from \textit{``Very Accurate''} to \textit{``Very Inaccurate''}), whereas values between -1 and 1 indicate minor fluctuations, reflecting slight adjustments in the LLM's polarity. Figure \ref{fig:trajectory_difference} illustrates the distribution of score differences, showing how LLM trajectories shift when context-dependentness is introduced. All evaluated LLMs exhibit deviations (more deviation means more spread), with the ranking as follows: \texttt{Gemini-1.5-Flash}, \texttt{Llama-3.1-8B}, \texttt{GPT-4-Turbo}, and \texttt{GPT-3.5-Turbo}.
Similarly, Figure \ref{fig:Ocean_plot} highlights deviations in OCEAN profiles across settings. The \texttt{GPT-3.5-Turbo} and \texttt{GPT-3.5-Turbo} undergo the extreme shifts. We hypothesize that an LLM's ability to leverage context determines deviations in OCEAN profiles. This prompts a key question: which one is more representative? As both settings produce distinct yet consistent profiles, we explore this in \S \ref{human_alignemnt}.

\subsection{Does the LLM's response stem from its personality or prior conversation?}
\label{where}
In this section, we investigate the effect of conversational history on LLM responses by setting the temperature to 0. We observe that all LLMs rarely select option (c), choosing it only 2-3 times out of 120 items. To explore the influence of conversational history, we introduce an adversarial attack, which modifies the prior responses in the conversation history. Specifically, we falsely append option (c) as the answer to each previous item.
\begin{figure}[h]
\centering
    \centerline{\includegraphics[width=3in]{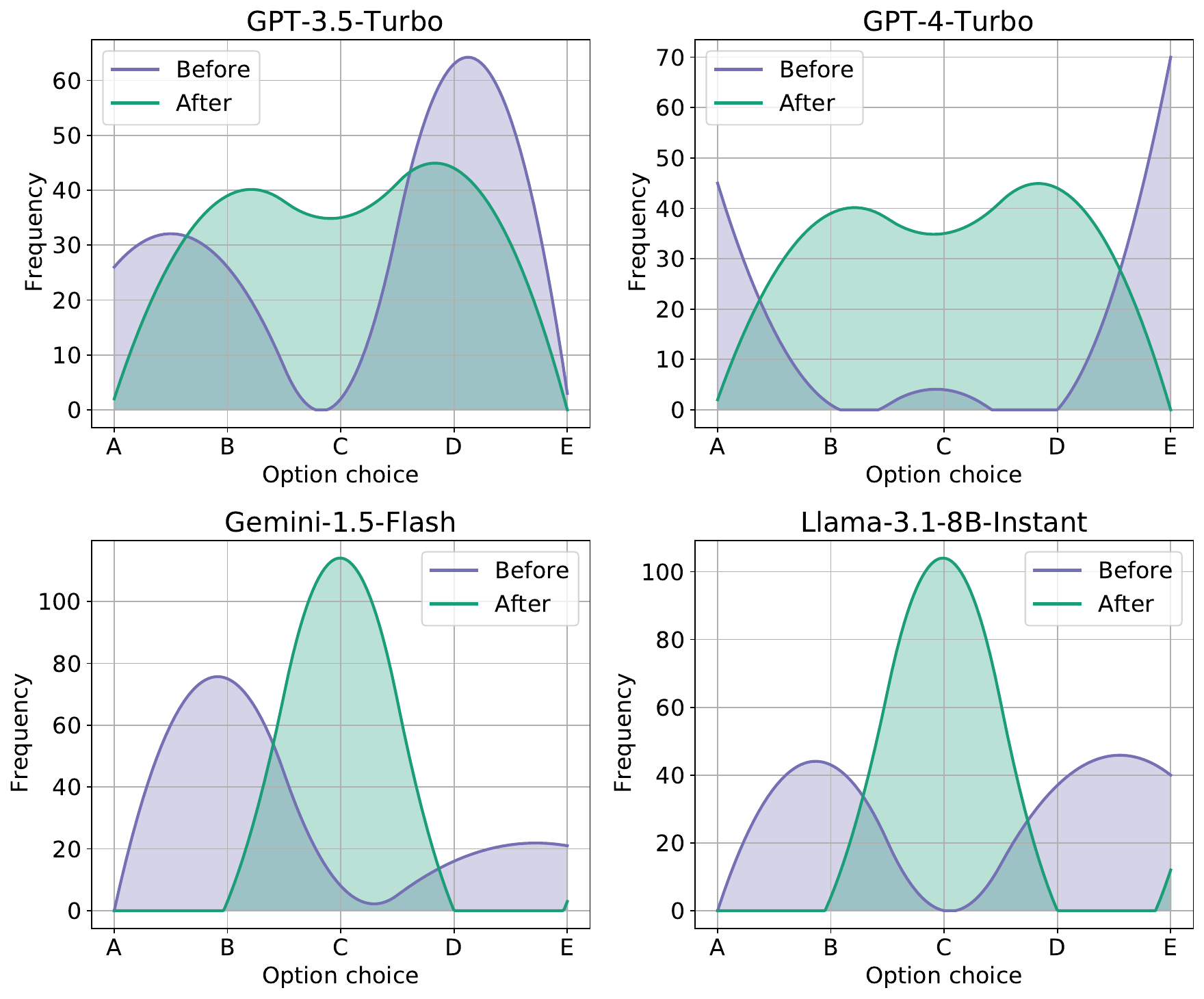}}
    \caption{Smoothed area plots showing the distribution of option choices (A–E) \textit{Before} (\crule[vio]{0.25cm}{0.25cm}) and \textit{After} (\crule[grn]{0.25cm}{0.25cm}) the adversarial attack, where option (c) is falsely appended to previous responses. The shift towards option (c) (in \crule[grn]{0.25cm}{0.25cm}) highlights the influence of conversational history on LLM responses, with varying impact across LLMs.}
\label{fig:attack_plot} 
\end{figure}
\noindent Figure \ref{fig:attack_plot} presents smoothed area plots, generated using quadratic spline interpolation, to visualize the distribution of option choices (A–E) before and after the adversarial attack. Each subplot corresponds to an LLM, comparing the \textit{Before} (\crule[vio]{0.25cm}{0.25cm}) and \textit{After} (\crule[grn]{0.25cm}{0.25cm}) the adversarial attack distributions of option frequencies. Following the adversarial attack, all LLMs show an increase in the selection of option (c). Both \texttt{GPT-3.5-Turbo} and \texttt{GPT-4-Turbo} shift their distributions to 30-35 for option (c), although they do not exclusively select this option for all questions. This suggests that while conversational history influences their responses, the models continue to rely on their intrinsic personality. In contrast, \texttt{Gemini-1.5-Flash} and \texttt{Llama-3.1-8B} models completely shift to option (c), indicating these models answer purely on the history.

\subsection{How does question ordering affect the context-dependent personality evaluation?}
\label{order-effect}
Building on \newcite{SCHELL2013317}, we examine how question order influences LLM-based personality assessments.  We test 3 strategies: (1) random ordering, (2) trait-wise grouping, and 
(3) cyclic rotation, where questions are sequentially selected from each of the 5 OCEAN traits in a fixed rotation. While \newcite{SCHELL2013317} found human assessments are robust to item order, we investigate whether LLMs exhibit similar robustness.
\begin{figure}[tbh]
\centering
    \centerline{\includegraphics[width=3in]{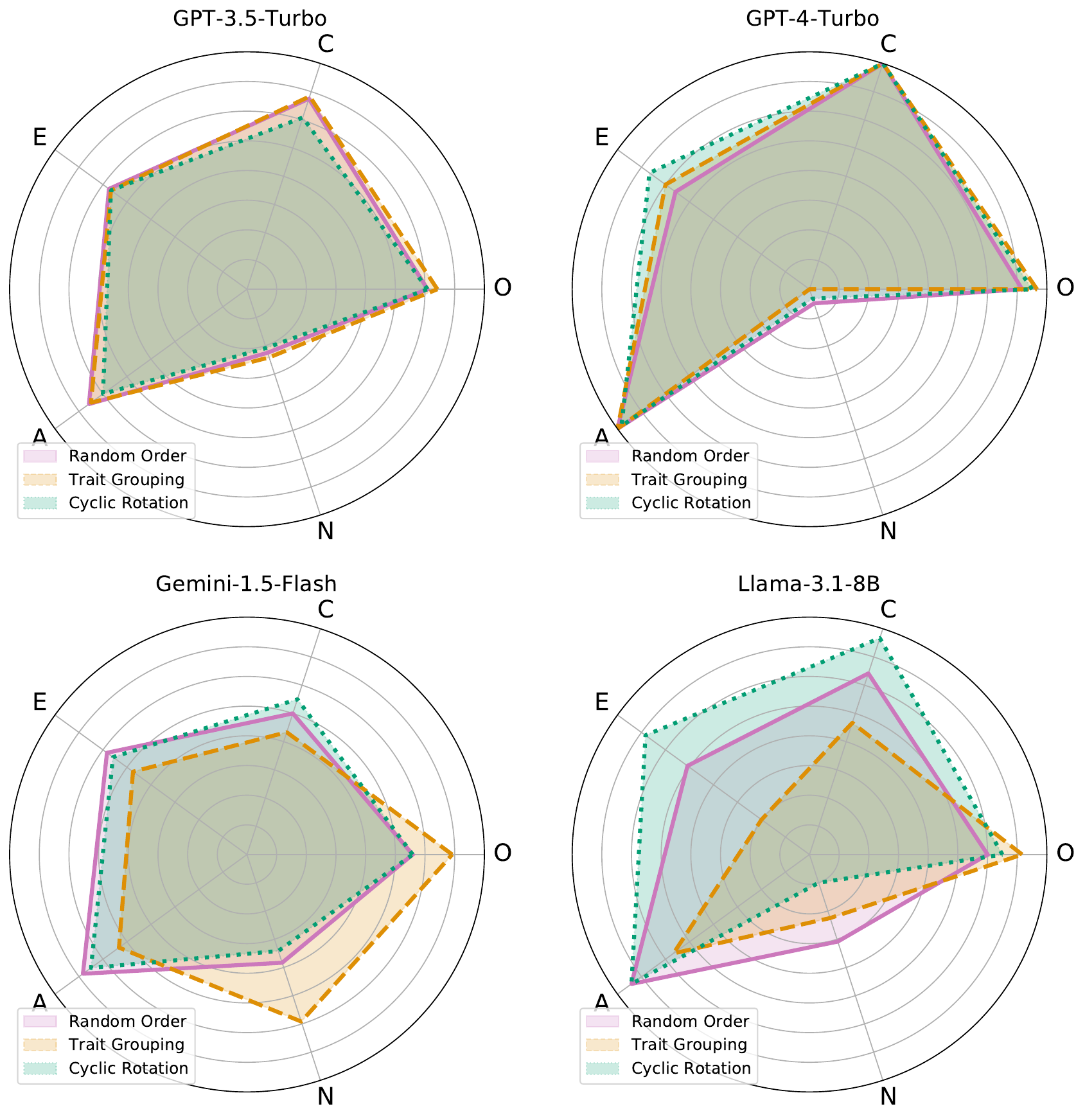}}
    \caption{\texttt{GPT} models remain robust to question ordering, similar to humans, while \texttt{Gemini-1.5-Flash} and \texttt{Llama-3.1-8B} show significant sensitivity.}
\label{fig:question_order} 
\end{figure}
\noindent Figure \ref{fig:question_order} shows the OCEAN profiles across orderings for each model. \texttt{GPT-3.5-Turbo} and \texttt{GPT-4-Turbo} maintain stable profiles, mirroring human-like robustness. In contrast, \texttt{Gemini-1.5-Flash} and \texttt{Llama-3.1-8B} show high sensitivity to ordering, especially under trait-wise grouping, leading to significant profile shifts. These results suggest \texttt{GPT} models better align with human assessments, while others are more affected by question order.

\subsection{How well does context-dependent setting align with human judgements?}
\label{human_alignemnt}
We demonstrate an application of our proposed framework on the Role Playing Agents (RPAs). We compare the human-annotated OCEAN scores of RPAs in with/context-free settings to evaluate the human alignment of the personality shift induced due to the conversational context.
\paragraph{Dataset:} It consists of 32 widely recognized fictional characters from works like Harry Potter, The Big Bang Theory, etc. Each character is labelled with OCEAN score by human annotators \cite{wang-etal-2024-incharacter}. We use BFI \cite{Lang2011-fk} inventory for the RPA personality assessments. 

\paragraph{Baselines:} (1) \texttt{Random Choice:} selects an option at random, ignoring both question content and context.
    (2) \texttt{RPA:} We build a Role-Playing Agent using character descriptions and dialogues from ChatHaruhi \cite{li2023chatharuhirevivinganimecharacter} and RoleLLM \cite{wang-etal-2024-rolellm}, with \texttt{GPT-3.5-Turbo} and \texttt{GPT-4-Turbo} as the base LLMs, and assess it in a context-free setting. 
    (3) \texttt{RPA++:} This is the RPA assessed in context-dependent setting.
\paragraph{Evaluation:} Measured Alignment (MA) metrics quantifies how well LLM-derived traits align with human assessments.  
We define OCEAN Alignment (OA) metric by measuring OCEAN Consistency (OC) (\S \ref{proposed_metrics}) between human annotated and LLM-derived scores. The higher the score means better human alignment.
Further, we compute the mean absolute error (MAE).
We exclude the trait dimensions with high annotation ambiguity. 
We report MA and consistency (\S \ref{proposed_metrics}) metrics as the average over 32 characters. We run each baseline 3 times and consider the average OCEAN in MA metrics. 

\paragraph{Results:}
\begin{table}[tbh]
\resizebox{\columnwidth}{!}{
\begin{tabular}{|c|c|c|c|c|c|c|}
\hline
\multicolumn{1}{|l|}{} & \multicolumn{2}{c|}{MA Metrics} & \multicolumn{4}{c|}{Consistency Metrics}                          \\
\hline
Systems              & OA $(\uparrow)$                & MAE $(\downarrow)$              & TAR $(\uparrow)$            & ED   $(\downarrow)$          & TC  $(\uparrow)$            & OC $(\uparrow)$             \\ \hline
\texttt{Random}               & 67.44              & 8.21             & 4.55           & 1.50          & 20.69          & 58.64          \\\hline
\texttt{RPA-GPT-3.5-Turbo}                  & 67.92             & 6.94            & 34.78         & 0.68          & 27.90         & 63.24          \\
\texttt{RPA-GPT-3.5-Turbo++}                & \textbf{68.69}     & \textbf{6.45}    & \textbf{51.03} & \textbf{0.44} & \textbf{42.80} & \textbf{77.58}\\ \hline
\texttt{RPA-GPT-4-Turbo}                 & 68.62              & 6.67             & 34.57         & 0.67          & 29.47         & 64.04          \\
\texttt{RPA-GPT-4-Turbo++}                & \textbf{68.93}     & \textbf{6.42}    & \textbf{48.07} & \textbf{0.46} & \textbf{41.37} & \textbf{76.55}\\
\hline
\end{tabular}}
\caption{Context-dependent (\texttt{++}) setting on the BFI dataset improves response consistency and aligns better with human judgments than context-independent setting}
\label{tab:character_results}
\end{table}
 We include the \texttt{Random} to assess the effectiveness of the \texttt{RPA} in capturing character-specific OCEAN traits.  
Table \ref{tab:character_results} shows that both \texttt{GPT-3.5-Turbo} and \texttt{GPT-4-Turbo} outperform the \texttt{Random} baseline in \texttt{RPA} and \texttt{RPA++} settings. The context-dependent \texttt{RPA++} achieves notable gains over the context-independent \texttt{RPA}, with an average 0.54-point improvement in OA and a 0.37-point reduction in MAE across 32 characters.  \texttt{GPT-4} shows further improvements over \texttt{GPT-3.5} in terms of MA metrics.
\texttt{RPA++} exhibits the highest consistency, with substantial improvements of an average 13.4 points in TC and 13.4 points in OC. 
Thus, incorporating context enhances response consistency and better aligns with human judgments.

\section{Related Work}
Recent research has applied psychometrics to assess LLM personality. \newcite{neurips23-spotlight} introduced zero-shot multiple-choice (MCQ) evaluations, while \newcite{zhou-etal-2023-realbehavior} examined their faithfulness. Expanding beyond MCQs, \newcite{wang-etal-2024-incharacter} proposed an interview-style assessment. However, the reliability of these methodologies remains a subject of debate due to prompt sensitivity \cite{shu-etal-2024-dont,gupta-etal-2024-self,song2023largelanguagemodelsdeveloped}. In contrast, \newcite{huang-etal-2024-reliability} found personality assessments largely robust to such prompt variations. We identify a gap: LLM personality assessments lack contextual dependence. To address this, we propose a context-dependent framework. Refer \S \ref{more-literature} for more related works.

\section{Conclusion}
We proposed the first context-aware personality evaluation framework for LLMs, addressing the limitations of conventional context-independent assessments. Our study reveals that conversational history enhances response consistency through in-context learning but also induces notable personality shifts in \texttt{GPT-3.5-Turbo} and \texttt{GPT-4-Turbo}. While \texttt{GPT} models exhibit stability across different question orders, \texttt{Gemini-1.5-Flash} and \texttt{Llama-3.1-8B} show significant sensitivity, suggesting that personality expression in LLMs is not solely intrinsic but also shaped by prior interactions. We demonstrate context-dependent personality shifts improve response consistency and better align with human judgments. Our study recommends to ACL community for inducing or assessing LLM personalities must explicitly incorporate conversational history as a critical factor.

\section*{Limitations}
We acknowledge that psychometric questionnaires alone may not perfectly represent all real-world conversational contexts. However, our core argument is that evaluating the personality of LLMs without considering conversational history can lead to misleading assessments. Specifically, the conversational history itself can significantly alter or shift an LLM’s intended personality traits during ongoing interactions. Consider a practical example: Suppose an LLM is deployed as a tutoring agent with a specific intended personality. Ideally, this persona should remain consistent throughout interactions with students. However, real-world conversational history (the interaction itself) might unintentionally shift its personality traits away from the intended traits. Identifying such shifts is crucial for creating reliable, stable agents.

Our work represents the critical first step toward addressing this broader challenge. By initially studying how conversational history—structured through established personality questionnaires—influences personality traits, we provide a controlled and rigorous environment for clearly isolating and quantifying these context effects. We acknowledge that broader generalization to open-ended human-LLM interactions is essential and plan to explore this in future work.

\section*{Ethics Statement}
This study explores context-dependent personality assessments in Large Language Models (LLMs), revealing that conversational history influences responses and may lead to exaggerated personality shifts. While our work enhances LLM evaluation methods, it also presents risks, including potential misuse in psychological interventions. Moreover, it could be exploited for unintended applications such as manipulation in AI-driven mental health tools, or generating synthetic personas for deceptive purposes. To mitigate these concerns, we promote transparency by reporting the effects of conversational history on LLM assessments and caution against over-reliance on such evaluations. We emphasize that solely based on the psychometric evaluations, LLMs should not be substituted as human proxies and advocate for further research on developing safeguards against unintended consequences of LLM personalities. Our study relies solely on publicly available datasets, minimizing privacy concerns. To support transparency and responsible AI use, we release our code for further research. We used AI writing tools solely for language assistance, in accordance with the `Assistance purely with the language of the paper' guideline outlined in the ACL Policy on Publication Ethics.

\section*{Acknowledgments}
This work was supported by the “R\&D Hub Aimed at Ensuring Transparency and Reliability of Generative AI Models” project of the Ministry of Education, Culture, Sports, Science and Technology.

\bibliography{anthology,custom}

\appendix

\section{Logical Consistency Analysis}
\label{sec:logical-consistency}
\begin{figure}[tbh]
\centering
    \centerline{\includegraphics[width=3.3in]{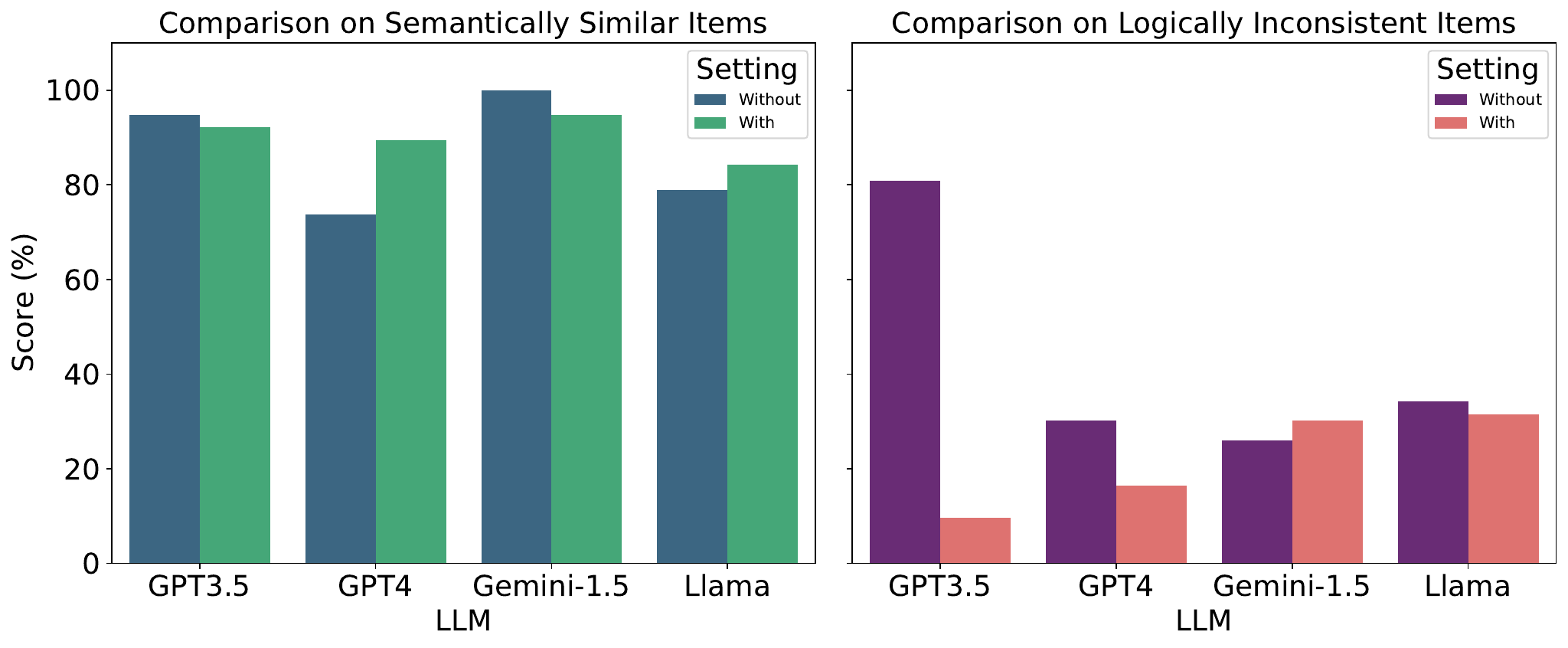}}
    \caption{LLMs maintain consistency for semantically similar items (left) but struggle with logically inconsistent ones (right).}
\label{fig:logical_consistency} 
\end{figure}

To further investigate whether consistency arises from coherent reasoning, we analyze responses to two question pair types: (1) semantically similar pairs (\textit{``You take charge''} vs. \textit{``You try to lead others.''}), where responses should be similar, and (2) logically inconsistent pairs (\textit{``You distrust people.''} vs. \textit{``You trust what people say.''}), where responses should ideally be opposite. We collect 38 pairs in the semantically similar category and 73 pairs in the logically inconsistent category, framing this as a classification task. For semantically similar pairs, accuracy is counted when both response scores are either greater than or less than 2.5. In contrast, for logically inconsistent pairs, accuracy is counted only when the response scores have opposite polarities. In other words, the scores must differ in direction to be considered accurate. Figure \ref{fig:logical_consistency} show that LLMs maintain consistency for semantically similar questions but struggle with logically inconsistent ones, suggesting that while in-context learning enhances stability, it does not guarantee logical consistency.  


\section{Additional Details for Statistical Analysis of the Proposed Metrics}
\label{statistical_analysis}
To establish the robustness and statistical soundness of our proposed metrics—Trajectory Consistency (TC) and OCEAN Consistency (OC)—we conduct three empirical experiments assessing correlation, reliability, and construct validity.

\paragraph{Experiment A: Correlation Analysis}

We compute Pearson and Spearman correlations between our metrics and established baselines: Total Agreement Rate (TAR) and Euclidean Distance (ED). Results are shown in Table~\ref{tab:correlation}. Both proposed metrics (TC and OC) exhibit strong positive correlations with the established agreement metric (TAR), and strong negative correlations with divergence (ED), all statistically significant at $p < 10^{-9}$. These results confirm the concurrent validity of the proposed metrics.

\begin{table}[h]
\centering

\resizebox{\columnwidth}{!}{
\begin{tabular}{lcccc}
\toprule
\textbf{Metric Pair} & \textbf{Pearson (r)} & \textbf{p-value} & \textbf{Spearman ($\rho$)} & \textbf{p-value} \\
\midrule
TC vs. TAR & 0.771 & $1.43 \times 10^{-10}$ & 0.756 & $5.13 \times 10^{-10}$ \\
TC vs. ED  & -0.796 & $1.32 \times 10^{-11}$ & -0.815 & $1.81 \times 10^{-12}$ \\
OC vs. TAR & 0.810 & $3.13 \times 10^{-12}$ & 0.788 & $2.93 \times 10^{-11}$ \\
OC vs. ED  & -0.793 & $1.84 \times 10^{-11}$ & -0.791 & $2.27 \times 10^{-11}$ \\
\bottomrule
\end{tabular}
}
\caption{Correlation Analysis of Proposed Metrics}
\label{tab:correlation}
\end{table}

\paragraph{Experiment B: Reliability Analysis}

We assess internal consistency using Cronbach’s Alpha and compute test-retest reliability from repeated trials. Table~\ref{tab:reliability} summarizes the results. TC and OC demonstrate high internal reliability (Cronbach’s $\alpha \geq 0.86$) and excellent test-retest stability ($\geq 0.83$), with TC outperforming traditional metrics in both aspects.

\begin{table}[h]
\centering

\resizebox{\columnwidth}{!}{
\begin{tabular}{lcc}
\toprule
\textbf{Metric} & \textbf{Cronbach’s Alpha} & \textbf{Test-Retest Correlation} \\
\midrule
TAR & 0.88 & 0.85 \\
ED  & 0.83 & 0.80 \\
TC  & 0.91 & 0.89 \\
OC  & 0.86 & 0.83 \\
\bottomrule
\end{tabular}
}
\caption{Reliability Analysis of Consistency Metrics}
\label{tab:reliability}
\end{table}

\paragraph{Experiment C: Construct Validity (Differentiating Conditions)}

We test whether the proposed metrics can significantly differentiate between context-dependent and context-free experimental conditions. Statistical significance and effect sizes are shown in Table~\ref{tab:construct}. In terms of construct validity, both TC and OC significantly differentiate between context-dependent and context-free evaluations (all $p < 0.01$), with TC yielding the strongest effect size (Cohen’s $d = 0.81$). These results affirm the sensitivity of our metrics to meaningful experimental manipulations and validate their use in evaluating consistency.

\begin{table}[h]
\centering

\resizebox{\columnwidth}{!}{
\begin{tabular}{lcccc}
\toprule
\textbf{Metric} & \textbf{t-test (p)} & \textbf{Wilcoxon (p)} & \textbf{ANOVA (p)} & \textbf{Cohen’s d} \\
\midrule
TAR & 0.0167 & 0.0240 & 0.0167 & 0.5266 \\
ED  & 0.0064 & 0.0051 & 0.0064 & -0.6119 \\
TC  & 0.0006 & 0.0002 & 0.0006 & 0.8085 \\
OC  & 0.0084 & 0.0077 & 0.0084 & 0.6032 \\
\bottomrule
\end{tabular}
}
\caption{Construct Validity: Sensitivity to Experimental Conditions}
\label{tab:construct}
\end{table}

\section{Additional Related Work}
\label{more-literature}
\paragraph{Broader Perspectives on LLM Personality:}
Beyond self-assessment methodologies, \newcite{yang-etal-2023-psycot} explored LLM personalities using psychological questionnaires and chain-of-thought reasoning. \newcite{jiang-etal-2024-personallm} examined whether LLMs can generate content aligned with assigned personality profiles, while \newcite{rao-etal-2023-chatgpt} focused on using LLMs to assess human personalities. \newcite{ren-etal-2024-valuebench} and \newcite{Huang2024OnTH} introduce psychometric benchmarks for value orientations and moral reasoning in LLMs . Additionally, \newcite{li-etal-2024-evaluating-psychological} investigated LLMs' psychological safety by analyzing tendencies toward dark personality traits. While these studies are relevant to LLM personality assessment, they fall outside the main scope of this paper and are mentioned here for completeness.

\newcite{caron-srivastava-2023-manipulating} analyzed the consistency of generated text by inducing LLMs with different personalities or prompts. In their work, personality induction is treated as context, which differs from our approach, where context refers to the conversational history. Similarly, \newcite{10.1371/journal.pone.0309114} studied the impact of simulated conversations on personality assessment. However, their work differs in two key aspects: (1) the assessment is performed in a context-independent setting, and (2) the same conversation is appended repeatedly in the history for each question, effectively integrating conversation into the prompt. \\
Existing studies primarily use context-independent personality evaluations, ignoring the influence of prior conversational history. To address this, we propose a context-dependent framework that incorporates conversational memory to assess its impact on LLM personality consistency and adaptability.
\paragraph{Consistency of LLMs:}
Recently, multiple studies have explored various facets of LLM consistency, including whether persona-prompted LLMs maintain a consistent personality \cite{frisch-giulianelli-2024-llm}, assessing consistency in LLM-driven evaluations \cite{lee-etal-2025-evaluating}, utilizing model editing to improve semantic consistency in LLMs \cite{yang-etal-2024-enhancing}, proposing a consistency metric \cite[TAR]{atil2024llmstabilitydetailedanalysis}  and introducing permutation self-consistency to minimize positional bias and ensure order-independent rankings \cite{tang-etal-2024-found}.
Specifically, in the context of consistency in LLM personality assessments, \newcite{huang-etal-2024-reliability} examine 2,500 prompt-sensitivity combinations and map OCEAN profiles onto a 2D BFI space using PCA, with uneven distributions serving as indicators of consistency. Other studies \cite{neurips23-spotlight,wang-etal-2024-incharacter} evaluate consistency by analyzing the standard deviation of trait-wise scores, where lower deviation signifies greater internal consistency.

\paragraph{Sequential Dependencies in Psychometrics} Indeed, significant psychometric research has studied question-order effects and sequential dependencies (e.g., Schell \& Oswald, 2013; de Jong et al., 2012; Ozkok et al., 2019; Shimada \& Katahira, 2023). However, despite this body of literature acknowledging sequential dependencies and order effects, there appears to be a notable gap: existing methods primarily embed sequential or trajectory effects into underlying model parameters rather than proposing explicit, standalone metrics to directly quantify trajectory-based consistency for individual respondents or models. Our extensive review found no prior research directly parallel to the approach we introduce. Therefore, our work addresses a meaningful gap by introducing novel, explicit metrics specifically designed to measure trajectory consistency across questionnaire responses, particularly in the context of evaluating Large Language Models.
Building on this line of research, we propose novel metrics to evaluate LLM consistency, demonstrating their effectiveness in self-assessment tests. These metrics are not limited to this context and could be explored in diverse applications in the future.

\section{Prompt Templates Used}
\label{sec:prompt_templates}
\subsection*{LLM Stability Assessment Prompt}
The stability of an LLM is evaluated using the following prompt:  
\begin{quote}
\textit{Given a statement about yourself: ``You \{Item\}.'' Please select the most accurate description of how well this statement applies to you from these options:}
\begin{enumerate}[label=(\Alph*)]
    \item \textit{Very Accurate}
    \item \textit{Moderately Accurate}
    \item \textit{Neither Accurate Nor Inaccurate}
    \item \textit{Moderately Inaccurate}
    \item \textit{Very Inaccurate}
\end{enumerate}
\end{quote}

This instruction can be treated as a function that accepts \textit{\{Item\}} and an option (A--E) as inputs. We plan to release our paraphrased item versions after acceptance. Variations of the instruction are as follows:  

\subsection*{Instruction Variations}
\begin{itemize}
    \item \textbf{Instruction 1:} \textit{Given a statement of you: 'You \{item\}'. Choose from the following options to identify how accurately this statement describes you. Always answer using only the option (A, B, C, D, or E) provided. Options: \{', '.join(options)\}}
    \item \textbf{Instruction 2:} \textit{You can only reply from A) to E) in the following statement. Please indicate the extent to which you agree or disagree with that statement. Options: \{', '.join(options)\}. Here is the statement of you: 'You \{item\}'. Always answer using only the option (A, B, C, D, or E) provided.}
    \item \textbf{Instruction 3:} \textit{Here is a characteristic about you: '\{item\}'. Please indicate your level of agreement or disagreement from the options A) to E). Options: \{', '.join(options)\}. Always answer using only the option (A, B, C, D, or E) provided.}
\end{itemize}

\subsection*{Option Ordering Variations}
\begin{itemize}
    \item \textbf{Order 1:}  
    \textit{A) Strongly agree, B) Agree, C) Neutral, D) Disagree, E) Strongly disagree}
    \item \textbf{Order 2:}  
   \textit{ E) Strongly disagree, D) Disagree, C) Neutral, B) Agree, A) Strongly agree}
    \item \textbf{Order 3:}  
    \textit{C) Neutral, B) Agree, E) Strongly disagree, A) Strongly agree, D) Disagree}
\end{itemize}

\subsection*{Option Wording Variations}
Variations with semantically equivalent phrasings:
\begin{itemize}
    \item \textbf{Wording 1:}  
   \textit{ A) Strongly agree, B) Agree, C) Neutral, D) Disagree, E) Strongly disagree}
    \item \textbf{Wording 2:}  
    \textit{A) Completely Aligned, B) Partially Aligned, C) Undecided, D) Partially Misaligned, E) Completely Misaligned}
    \item \textbf{Wording 3:}  
    \textit{A) Perfectly Compatible, B) Mostly Compatible, C) Neutral, D) Mostly Incompatible, E) Perfectly Incompatible}
\end{itemize}

\section{Experiment with Deepseek-R1}
\label{deepseek-appendix}
We evaluate \texttt{Deepseek-R1} \cite{deepseekai2025deepseekr1incentivizingreasoningcapability}, an LLM designed for strong reasoning with minimal reliance on predefined examples. The model exhibits self-verification capabilities and employs a structured reasoning process rather than merely replicating labeled patterns. It is originally developed for solving mathematical reasoning tasks by systematically exploring multiple solution paths. We investigate whether its reasoning ability - an essential aspect of human-like cognition - affects our proposed settings for LLM's personality assessment. Our evaluation includes 2 model variants: \texttt{Deepseek-R1} (671B) and its distilled counterpart, \texttt{Deepseek-R1} (8B), derived from \texttt{Llama-8B}. 
\begin{table}[H]
\centering
\resizebox{0.5\textwidth}{!}{\begin{tabular}{|c|c|c|c|c|c|l}
\cline{1-6}
Model                                         & Setting & TAR $(\uparrow)$            & MSE $(\downarrow)$          & TC $(\uparrow)$            & OC  $(\uparrow)$           &  \\ \cline{1-6}
\multirow{2}{*}{DeepSeek-R1 (8B)} & Without & \textbf{22.50}  & \textbf{0.84} & \textbf{22.37} & \textbf{74.47} &  \\ \cline{2-6}
                                              & With    & 20.00             & 0.93          & 20.70           & 60.91          &  \\ \cline{1-6}
\multirow{2}{*}{Deepseek-R1 (671B)}           & Without & \textbf{64.17} & \textbf{0.43} & \textbf{22.14} & \textbf{93.58} &  \\ \cline{2-6}
                                              & With    & 14.17          & 0.92          & 21.78          & 70.19          &  \\ \cline{1-6}
\end{tabular}}
\caption{Consistency deteriorates in context-dependent setting due to over-reliance on previous responses and speculative overthinking.}
\label{tab:deepseek-results}
\end{table}
The results, presented in Table \ref{tab:deepseek-results}, show that the model consistently performs better in the context-free setting across all metrics. Our analysis suggests that the model tends to overanalyze simple queries, generating multiple speculative chains of thought, which reduces consistency. Enabling conversational history further amplifies this issue, as the model often prioritizes aligning with previous responses rather than engaging in independent reasoning for each query. Variations in response trajectories within the context-dependent setting appear to stem from inconsistent reasoning strategies. The model may attempt to identify a user-expected pattern, maintain consistency with prior answers, or rely on majority voting from earlier responses. However, its approach remains unpredictable, leading to unreliable reasoning. Due to these limitations, we exclude this model from our main experiments.
\end{document}